\documentclass[10pt,twocolumn,letterpaper]{article}

\usepackage{wacv}
\usepackage{times}
\usepackage{epsfig}
\usepackage{graphicx}
\usepackage{amsmath}
\usepackage{amssymb}
\usepackage{cite}
\usepackage{hyperref}
\usepackage{multirow}
\usepackage{wrapfig}
\usepackage[numbers]{natbib}

\wacvfinalcopy 

\def\wacvPaperID{****} 

\begin{document}

\title{Similarity Learning Networks for Animal Individual Re-Identification - Beyond the Capabilities of a Human Observer}

\author{Stefan Schneider \\
\small{School of Computer Science,} \\ 
\small{University of Guelph}\\
{\tt\small sschne01@uoguelph.ca}
\and
Graham W. Taylor \\
\small{School of Engineering,} \\
\small{University of Guelph} \\ 
\small{Vector Institute for} \\ 
\small{Artificial Intelligence} \\
{\tt\small gwtaylor@uoguelph.ca}
\and
Stefan C. ~Kremer \\
\small{School of Computer Science,} \\ 
\small{University of Guelph} \\
{\tt\small skremer@uoguelph.ca}}

\maketitle

\begin{abstract}
Deep learning has become the standard methodology to approach computer vision tasks when large amounts of labeled data are available. One area where traditional deep learning approaches fail to perform is one-shot learning tasks where a model must correctly classify a new category after seeing only one example. One such domain is animal re-identification, an application of computer vision which can be used globally as a method to automate species population estimates from camera trap images. Our work demonstrates both the application of similarity comparison networks to animal re-identification, as well as the capabilities of deep convolutional neural networks to generalize across domains. Few studies have considered animal re-identification methods across species. Here, we compare two similarity comparison methodologies: Siamese and Triplet-Loss, based on the AlexNet, VGG-19, DenseNet201, MobileNetV2, and InceptionV3 architectures considering mean average precision (mAP)@1 and mAP@5. We consider five data sets corresponding to five different species: humans, chimpanzees, humpback whales, fruit flies, and Siberian tigers, each with their own unique set of challenges. We demonstrate that Triplet Loss outperformed its Siamese counterpart for all species. Without any species-specific modifications, our results demonstrate that similarity comparison networks can reach a performance level beyond that of humans for the task of animal re-identification. The ability for researchers to re-identify an animal individual upon re-encounter is fundamental for addressing a broad range of questions in the study of population dynamics and community/behavioural ecology. Our expectation is that similarity comparison networks are the beginning of a major trend that could stand to revolutionize animal re-identification from camera trap data.
\end{abstract}
\section{Introduction}
Recent decades have witnessed the emergence of deep learning systems that make use of large data volumes \cite{zheng2015scalable}. Modern deep learning systems no longer require `hard-coded' feature extraction methods. Instead, these algorithms can learn, through their exposure to many examples, the particular features that allow for the discrimination of individuals. \cite{lecun2015deep}. Deep learning methods have shown great success when considering computer vision re-ID tasks with large amounts of data. However, for the task of animal re-ID, gathering a library of images for every individual within a population is infeasible.

Similarity comparison networks, such as Siamese and Triplet-Loss networks, are a popular alternative to standard deep networks and have shown success re-identifying (re-ID) human individuals \cite{schroff2015facenet}. Rather than traditional softmax classification outputs, Siamese networks consider pairs of inputs and classify them as either similar or dissimilar. For re-ID, this extends to determining if two input images are of the same individual. We believe the capabilities of these systems can extend beyond those of humans. To benchmark similarity comparison networks and understand their limitations, we compare such systems in a domain that has received little focus: non-human animals. Animal species provide an excellent test bed for the capabilities of similarity comparison networks as the characteristics that distinguish animal individuals are often much more subtle than that of humans. Here we explore five architectural variants of the Siamese paradigm: AlexNet, VGG-19, DenseNet201, MobileNetV2, and InceptionV3 to test their ability to re-ID individuals of five species: humans, chimpanzees (\textit{Pan spp.}), humpback whales (\textit{Megaptera novaeangliae}), fruit flies (\textit{Drosophila melanogaster}), and Siberian tigers (\textit{Panthera tigris altaica})\cite{krizhevsky2012imagenet, simonyan2014very, huang2017densely, sandler2018mobilenetv2, szegedy2016rethinking}.

Current practice requires years of training and practical experience. Ecologists rely on a variety of techniques for re-ID including: tagging, scarring, banding, and DNA analyses of hair follicles or feces \cite{krebs1989ecological}. While accurate, these techniques are laborious for the field research team, intrusive to the animal, and often expensive for the researcher. 

Re-identification from camera trap images is a desirable alternative for ecologists due to its lower cost and reduced workload for field researchers. Despite its advantages, there are a number of practical and methodological challenges associated with its use. Primarily, even among experienced researchers, there remains an opportunity for human error and bias \cite{foster2012critique, meek2013reliability}. Historically, these limitations have restricted the use of camera traps to the re-ID of animals that bear conspicuous individual markings \cite{foster2012critique}. 

Our objective is to test the generalization of similarity comparison networks considering re-ID tasks across multiple species as well as whether a deep learning system can expedite and reduce the human biases inherent to the task of re-identifying animals from camera trap images. Animal re-ID is used for a variety of ecological metrics, including diversity, relative abundance distribution, and carrying capacity \cite{krebs1989ecological}. By training an animal re-ID system, one could automate the collection of metrics relevant to health projection of ecosystem stability and species population health.

\section{Brief History of Computer Vision Methods for Animal Re-Identification}

Prior to the advent of deep learning, for decades, the approach to standardizing the statistical analysis of animal re-ID has involved computer vision. `Feature engineering' has been the most commonly used act of \emph{engineering as programming} where algorithms are designed and implemented to focus exclusively on predetermined traits, such as the detection of patterns of spots or stripes, to discriminate among individuals. The main limitations of this approach surround its impracticality \cite{hiby2009tiger}. Feature engineering requires programming experience, sufficient familiarity with the organisms to identify relevant features, and lacks in generality where once a feature detection algorithm has been designed for one species, it is unlikely to be useful for other taxa. For a comprehensive review of computer vision relevant to animal re-ID see  \citet{schneider2018deep}.

The success of deep learning methods for human re-identification is well documented when ample training images are available for each individual. In 2015, using standard convolutional architectures, two research teams, \citet{lisanti2015person} and \citet{martinel2015re} demonstrated the success of CNNs on human re-ID using ETHZ, a data set composed of 8580 images of 148 unique individuals taken from mobile platforms. CNNs were able to correctly classify individuals after seeing 5 images of an individual. \citet{taigman2014deepface} introduced Deepface, a method of creating a 3-dimensional representation of the human face to provide more data to a neural network which improved classification accuracy on the YouTube faces data set containing videos of 1,595 individuals. 

\begin{figure}[t]
\begin{center}
   \includegraphics[width=0.8\linewidth]{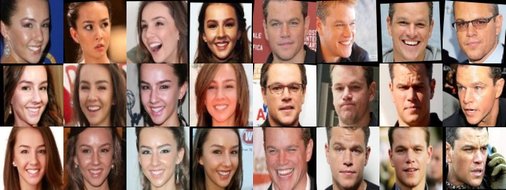}
\end{center}
   \caption{Example Images from the FaceScrub Data Set \cite{ng2014data}}
\label{fig:long}
\label{fig:onecol}
\end{figure}

Despite the success of deep learning methods for human re-ID, few ecological studies have realized its advantages. \citet{carter2014automated} published one of the first works using neural networks for animal re-ID, a tool for green turtle (\textit{Chelonia mydas}) re-ID. The authors collected 180 photos of 72 individuals from Lady Elliot Island in the southern Great Barrier Reef, both nesting and free swimming. They considered an undisclosed number of testing images. Their algorithm pre-processes the image by extracting a shell pattern, converting it to grey scale, unravelling the data into a raw input vector, and then training a simple feedforward network. Each individual model yields an output accuracy of 80-85\%, but the authors utilize an ensemble approach by training 50 different networks and having each vote for a correct classification. The ensemble approach attains an accuracy of 95\%. Carter et al.'s work has been considered a large success and is currently used to monitor the southern Great Barrier Reef green turtle population. 

\citet{freytag2016chimpanzee} trained the CNN architecture AlexNet on the isolated faces of chimpanzees considering two chimpanzee data sets: C-Zoo and C-Tai. They report an improved accuracy of 92.0\% and 75.7\% in comparison to the original Support Vector Machine method of 84.0\% and 68.8\% \cite{freytag2016chimpanzee, loos2013automated}. \citet{brust2017towards} trained the object detection method YOLO to extract cropped images of Gorilla (\textit{Gorilla gorilla}) faces from 2,500 annotated camera trap images of 482 individuals taken in the Western Lowlands of the Nouabal\'e -Nodki National Park in the Republic of Congo. Once the faces are extracted, \citet{brust2017towards} followed the same procedure as \citet{freytag2016chimpanzee} to train AlexNet, achieving a 90.8\% accuracy on a test set of 500 images. The authors herald the promise of deep learning for ecological studies show promise for a whole realm of new applications in the fields of basic identify, spatio-temporal coverage and socio-ecological insights.

\section{Similarity Learning Networks for Animal Re-Identification}

When approaching the problem of animal re-ID, traditional CNN architectures require a data set containing a large number of examples for every individual from the population. This is infeasible for real-world scenarios. Furthermore, they also require fixing the number of individuals in advance, so one cannot add individuals to the population without retraining the model. In order to utilize deep learning for animal re-ID, an alternative approach must be considered. 

\citet{bromley1994signature} introduced a suitable neural network architecture for this problem, the Siamese network, which learns to detect if two input images are similar or dissimilar \cite{bromley1994signature}. This approach allows new individuals to be recognized without example images of every individual in a population, and without any re-training of the network. Similarity comparison networks, such as the Siamese Network and Triplet Loss network, are forms of distant metric learning, which function by comparing the euclidean distance of a latent space embeddings, often at the last layer, considering two sister networks. This is in contrast to traditional network architectures which require all subjects to be identified and well represented in the training data and re-trained if a new individual were added to the data. The similarity comparison approach instead trains a network to learn how to identify similarities between two subjects. This allows new subjects to be recognized without example images of every individual in a population, and without any re-training of the network. Once trained, similarity comparison networks require only one labeled input image of an individual in order to accurately re-identify the second input image of the same individual. The main advantage for re-ID is that these systems generalize to individuals not found in the training data. For humans, \citet{schroff2015facenet} demonstrated the Triplet-Loss similarity comparison framework, FaceNet, which currently holds the highest accuracy on the YouTube Faces data set with a 95.12\% top-1 accuracy and is a promising model for animal re-ID. 

\begin{figure}[t]
\begin{center}
   \includegraphics[width=0.8\linewidth]{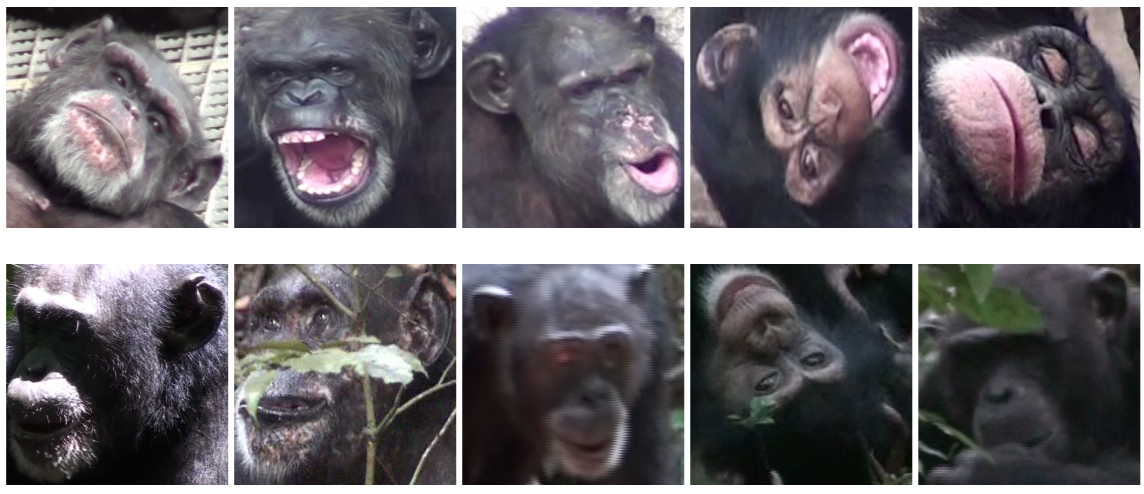}
\end{center}
   \caption{Example Images from the Chimpface Data Set \cite{freytag2016chimpanzee}}
\label{fig:long}
\label{fig:onecol}
\end{figure}

\citet{deb2018face} utilized Siamese networks for animal re-ID considering three species: chimpanzees, lemurs, and golden monkeys. They formulated the problem by defining three categories for testing successful re-ID: verification (determine if two images are the same individual), closed-set identification (identify an individual from a given set of images), and open-set identification (identify an individual from a given set of images or conclude the individual is absent from the data). For chimpanzees, they combined the C-Zoo and C-Tai data sets to create the \textit{ChimpFace} data set which contains 5,599 images of 90 chimpanzees. For lemurs, they consider a data set known as \textit{LemurFace} from the Duke Lemur Center, North Carolina which contains 3,000 face images of 129 lemur individuals from 12 different species. For golden monkeys, they extracted the faces of 241 short video clips (average 6 seconds) from Volcanoes National Park in Rwanda where 1,450 images of 49 golden monkey faces were cropped and extracted \cite{deb2018face}. They use a custom Siamese CNN containing four convolutional layers, followed by a 512 node fully connected layer \cite{deb2018face}. Deb et al. (2018) report the above defined verification, closed-set, and open-set accuracies respectively for lemurs: 83.1\%, 93.8\%, 81.3\%, golden monkeys: 78.7\%, 90.4\%, 66.1\%, and chimpanzees: 59.9\%, 75.8\%, and 37.1\%.

Triplet Loss networks have been found to outperform Siamese networks for the task of digit and human face recognition \cite{hoffer2015deep}. Triplet Loss networks differ from Siamese networks by maximizing the distance of the embedding between two pair wise images per sample: an anchor and positive pair, and an anchor and negative pair. One advantage of Triplet Loss networks is the ability to consider optimal image pairings per mini-batch \cite{schroff2015facenet}. This allows for a form of curriculum learning, where easy pairwise samples are selected early in training, and difficulty increases as validation loss decreases \cite{hermans2017defense}. In 2019, \citet{bouma2018individual} trained triplet loss networks for animal re-ID using images of dolphin fins. By following the describe triplet loss for the euclidean distance of positive and negative pairs, the learned embeddings were able to achieve 90.5\% top-1 accuracy for dolphins considering 37 test individuals. 

To date, no one has directly compared Siamese and Triplet Loss methodologies for animal re-ID across multiple data sets. We test their performances here.

\section{Methods}

To benchmark similarity networks on animal re-ID, we consider the verification accuracy metric proposed by \citet{deb2018face} on five species using the following data sets, each with their own unique challenges:

\begin{itemize}
	\item FaceScrub: 106,863 images of 530 male/female human individuals varying in pose \cite{ng2014data}. This data set allows for a benchmark comparison of our methodology in comparison to other human similarity networks.
	
	\item ChimpFace: 5,599 images of 95 male/female chimpanzee (\textit{Pan troglodytes}) individuals. This is a combination of two previous data sets: C-Tai and C-Zoo \cite{freytag2016chimpanzee}. This data set provides the unique opportunity of comparing the performance of similarity networks to the previously reported performance of feature engineering as well as classical deep learning methods.
	
	\item HappyWhale: 9,850 images of 4,251 humpback whale (\textit{Megaptera novaeangliae}) individuals offered as an expired Kaggle competition. This data set provides a realistic representation of the real-world application of animal re-ID as the 9,046 images only contain the fluke of the whale and are extremely sparse, having only an average of only 2 (+/- 8) individuals considering 4,251 individual classifications \cite{kagglehumpbackreid}.
	
	\item FruitFly: 244,760 images of 20 fruit flies (\textit{Drosophila melanogaster}) in a variety of poses \cite{schneider2018can}. Allows for the ability to test the capabilities of similarity learning networks on an animal species beyond the Chordata phylum, and in the Arthopoda phylum, where re-ID is beyond the capabilities of a human observer.
	
	\item Amur Tiger Re-identification in the Wild (ATRW): 1,870 images of 92 Siberian tigers (\textit{Panthera tigris altaica}). A recent effort to provide a large scale tiger re-ID data set \cite{li2019amur}. Tigers are captured in a diverse set of unconstrained poses and lighting conditions.
\end{itemize}	

One trend that is prevalent throughout the history of animal re-ID is the limitation of data for individual animal re-ID, especially those publicly available. From the data sets we selected, the Chimpface, FruitFly and the ATRW data sets are limited in the numbers of individuals typical for re-ID in comparison to human data sets and benchmarks \cite{schroff2015facenet}. This is an unfortunate reality of working with animal re-ID. To account for this, we divide our data using a 0.7/0.1/0.2 train/validation/test for the FaceScrub, Chimpface, HappyWhale, and ATRW data sets to increase the number of individuals. For the FruitFly data set we use a 0.4/0.1/0.5 split to increase the number of testing individuals so that a mAP@5 is a meaningful metric. In addition, for each experiment we perform a five-fold train/validation/testing split, providing an inferred re-ID capability for all individuals within the data set. 

\begin{table}[h!]
\centering
\caption{Summary of Data Splitting}
\begin{tabular}{ l c c c}
	\hline
	\multicolumn{1}{p{0.5cm}}{Species} &
	\multicolumn{1}{p{1.5cm}}{\centering Ratio} &
	\multicolumn{1}{p{1.5cm}}{\centering Num \\Individuals} & 
	\multicolumn{1}{p{1.5cm}}{\centering Num Images} \\
		\hline
		\multirow{3}{*}{FaceScrub} \\
	    & 0.7 & 414 & 73,735 \\
		& 0.1 & 52 & 10,152 \\
		& 0.2 & 52 & 22,976 \\
		\hline
		\multirow{3}{*}{ChimpFace} \\
	    & 0.7 & 67 & 3,891 \\
		& 0.1 & 10 & 588 \\
		& 0.2 & 19 & 1,120 \\
		\hline
		\multirow{3}{*}{HappyWhale} \\
	    & 0.7 & 2,978 & 6,914 \\
		& 0.1 & 425 & 985 \\
		& 0.2 & 850 & 1,951 \\
		\hline		
		\multirow{3}{*}{FruitFly} \\
	    & 0.4 & 8 & 97,904 \\
		& 0.1 & 2 & 24,476 \\
		& 0.5 & 10 & 122,380 \\
		\hline	
		\multirow{3}{*}{ATWR} \\
	    & 0.7 & 64 & 1467 \\
		& 0.1 & 9 & 163 \\
		& 0.2 & 18 & 250 \\
		\hline	
\end{tabular}
\end{table}

\begin{figure}[t]
\begin{center}
   \includegraphics[width=0.8\linewidth]{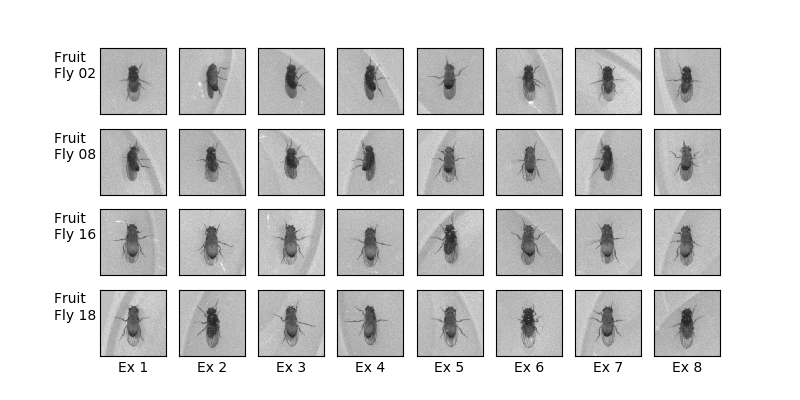}
\end{center}
   \caption{Example Images from the Fruit Fly Data Set \cite{schneider2018can}}
\label{fig:long}
\label{fig:onecol}
\end{figure}

\subsection{Performance, architecture and training details}

As a measure of performance we report the closed-set mAP@1 and mAP@5 re-ID accuracies. To calculate these values we use a sampling strategy where, for each individual in the testing set a random sample image is selected. One positive pair is randomly selected, as well as negative pairs for each remaining individual in the testing set. mAP@1 and mAP@5 are calculated considering the area under a precision/recall curve considering the top and top-5 model outputs respectively. This process is repeated 1000 times. This sampling approach is necessary due to the computational expense of an exhaustive search of all pairwise interactions, especially for the FruitFly data set.  

Considering the models themselves, we compare five standardized computer vision architectures: AlexNet, VGG19, DenseNet201, ResNet152, and InceptionNet V3 considering the two described similarity comparison methodologies: Siamese and Triplet-Loss. For Siamese, we consider the contrastive loss \cite{hadsell2006dimensionality}.

When initializing we use weights pre-trained on ImageNet. We also apply data augmentation methods during training using the torchvision library \cite{torchvision}. For each training example each augmentation strategy is randomly applied from a choice of: mirroring, shifting, rotation, colour channel noise to be added, per pixel manipulation, blurriness, and pixel dropout. 

For training we selected the Adaptive Momentum (Adam) optimizer with a learning rate of 0.001  and a decay rate of 2\% \cite{kingma2014adam}. To represent error, we consider the Siamese loss and Triplet Loss by minimizing the euclidean distance of the 128 embedding considering a margin size of 1 and utilize a semi-hard pairwise image selection strategy. The model was trained using mini-batch sizes of 32 training examples of size 224 x 244 pixel, except inception which was 299 x 299, and trained for 100 epochs. Training and analyses of this model were performed using Python 3.6 and Pytorch 1.1 on a NVIDIA P100 GPU.  

\begin{figure}[t]
\begin{center}
   \includegraphics[width=0.645\linewidth]{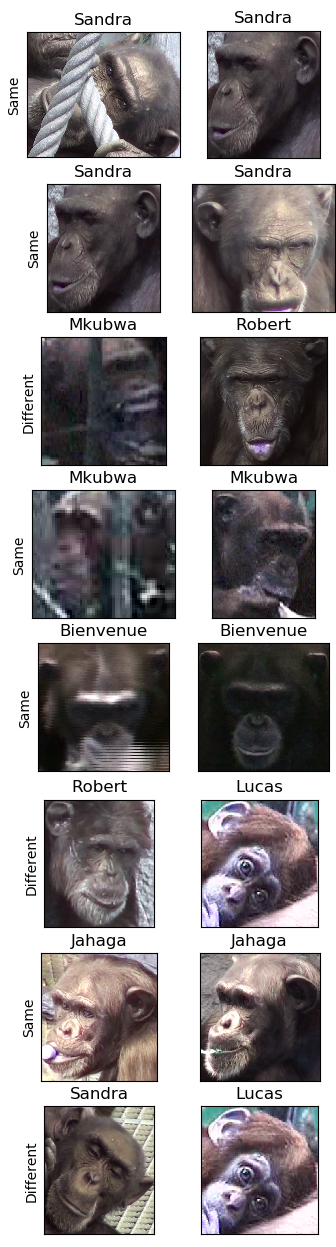}
\end{center}
   \caption{Model Output for Chimpanzee Individuals. Y label for each image is the model output and the X label is the name of each individual model's prediction \cite{freytag2016chimpanzee}}
\label{fig:long}
\label{fig:onecol}
\end{figure}

\section{Animal Re-Identification Results}

We find that for all species, triplet loss outperforms its Siamese network counterpart. We also found that for all mAP@5 scores, ResNet152 was the top performing model. This is in contrast to the mAP@1 scores, which Densnet201, ResNet152, and InceptionV3 performed optimally for different data sets. AlexNet and VGG19 we able to learn features relative to the task, but never outperformed the three previously mentioned architectures.

\begin{table}[h!]
\centering
\caption{Summary of Performance Metrics for Siamese Similarity Learning Models by Species \& Data Set}
\footnotesize
\begin{tabular}{ l c c c}
	\multicolumn{4}{ c }{}\\
	\hline
	\multicolumn{1}{p{1cm}}{Species} &
	\multicolumn{1}{p{1.5cm}}{\centering Model} &
	\multicolumn{1}{p{1.5cm}}{\centering mAP@1} & 
	\multicolumn{1}{p{1.5cm}}{\centering mAP@5} \\ \hline	
	\multirow{5}{*}{Human} 
	    & AlexNet & 0.699 $\pm$ 0.342 & 0.721 $\pm$ 0.332\\
		& VGG19 & 0.680 $\pm$ 0.288 & 0.703 $\pm$ 0.251\\
		& DenseNet201 & 0.734 $\pm$ 0.277 & 0.835 $\pm$ 0.875 \\
		& ResNet152 & \textbf{0.756 $\pm$ 0.282} & \textbf{0.856 $\pm$ 0.123}\\
		& InceptionV3 & 0.713 $\pm$ 0.227 & 0.854 $\pm$ 0.126\\
		\hline
	\multirow{5}{*}{Chimpanzee} 
	    & AlexNet & 0.639 $\pm$ 0.221 & 0.863 $\pm$ 0.121 \\
		& VGG19 & 0.645 $\pm$ 0.168 & 0.884 $\pm$ 0.094\\
		& DenseNet201 & 0.725 $\pm$ 0.134 & 0.871 $\pm$ 0.064 \\
		& ResNet152 & \textbf{0.775 $\pm$ 0.134} & \textbf{0.901 $\pm$ 0.097}\\
		& InceptionV3 & 0.743 $\pm$ 0.106 & 0.869 $\pm$ 0.125 \\
		\hline
	\multirow{5}{*}{Whale} 
	    & AlexNet & 0.509 $\pm$ 0.385 & 0.662 $\pm$ 0.334 \\
		& VGG19 & 0.543 $\pm$ 0.397 & 0.669 $\pm$ 0.410 \\
		& DenseNet201 & 0.521 $\pm$ 0.445 & 0.691 $\pm$ 0.312 \\
		& ResNet152 & 0.563 $\pm$ 0.202 & \textbf{0.737 $\pm$ 0.298} \\
		& InceptionV3 & \textbf{0.576 $\pm$ 0.203} & 0.722 $\pm$ 0.390 \\
		\hline
	\multirow{5}{*}{Fruit Fly} 
	    & AlexNet & 0.621 $\pm$ 0.078 & 0.875 $\pm$ 0.064\\
		& VGG19 & 0.590 $\pm$ 0.081 & 0.838 $\pm$ 0.090\\
		& DenseNet201 & 0.638 $\pm$ 0.153 & 0.843 $\pm$ 0.180\\
		& ResNet152 & \textbf{0.693 $\pm$ 0.098} & \textbf{0.896 $\pm$ 0.109} \\
		& InceptionV3 & 0.522 $\pm$ 0.021 & 0.873 $\pm$ 0.143 \\
		\hline	
	\multirow{5}{*}{Tiger} 
	    & AlexNet & 0.794 $\pm$ 0.396 & 0.858 $\pm$ 0.289\\
		& VGG19 & 0.735 $\pm$ 0.245 & 0.821 $\pm$ 0.243 \\
		& DenseNet201 & \textbf{0.803 $\pm$ 0.398} & 0.8756 $\pm$ 0.148 \\
		& ResNet152 & 0.789 $\pm$ 0.320 & \textbf{0.877 $\pm$ 0.172} \\
		& InceptionV3 & 0.701 $\pm$ 0.307 & 0.843 $\pm$ 0.231\\
		\hline	
\end{tabular}
\end{table}

\begin{table}[h!]
\centering
\caption{Summary of Performance Metrics for Triplet Loss Similarity Learning Models by Species \& Data Set}
\footnotesize
\begin{tabular}{ l c c c}
	\multicolumn{4}{ c }{}\\
	\hline
	\multicolumn{1}{p{1cm}}{Species} &
	\multicolumn{1}{p{1.5cm}}{\centering Model} &
	\multicolumn{1}{p{1.5cm}}{\centering mAP@1} & 
	\multicolumn{1}{p{1.5cm}}{\centering mAP@5} \\ \hline	
	\multirow{5}{*}{Human} 
	    & AlexNet & 0.739 $\pm$ 0.284 & 0.804 $\pm$ 0.345\\
		& VGG19 & 0.811 $\pm$ 0.325 & 0.843 $\pm$ 0.251\\
		& DenseNet201 & \textbf{0.914 $\pm$ 0.299} & 0.947 $\pm$ 0.187 \\
		& ResNet152 & 0.886 $\pm$ 0.301 & \textbf{0.952 $\pm$ 0.093}\\
		& InceptionV3 & 0.903 $\pm$ 0.235 & 0.940 $\pm$ 0.124\\
		\hline
	\multirow{5}{*}{Chimpanzee} 
	    & AlexNet & 0.739 $\pm$ 0.241 & 0.886 $\pm$ 0.166 \\
		& VGG19 & 0.734 $\pm$ 0.188 & 0.890 $\pm$ 0.085\\
		& DenseNet201 & 0.792 $\pm$ 0.164 & 0.932 $\pm$ 0.049 \\
		& ResNet152 & \textbf{0.811 $\pm$ 0.155} & \textbf{0.961 $\pm$ 0.097}\\
		& InceptionV3 & 0.756 $\pm$ 0.136 & 0.940 $\pm$ 0.075 \\
		\hline
	\multirow{5}{*}{Whale} 
	    & AlexNet & 0.679 $\pm$ 0.374 & 0.752 $\pm$ 0.274 \\
		& VGG19 & 0.713 $\pm$ 0.349 & 0.801 $\pm$ 0.287 \\
		& DenseNet201 & 0.691 $\pm$ 0.304 & 0.771 $\pm$ 0.253 \\
		& ResNet152 & 0.733 $\pm$ 0.252 & \textbf{0.830 $\pm$ 0.275} \\
		& InceptionV3 & \textbf{0.746 $\pm$ 0.243} & 0.804 $\pm$ 0.290 \\
		\hline
	\multirow{5}{*}{Fruit Fly} 
	    & AlexNet & 0.671 $\pm$ 0.158 & 0.935 $\pm$ 0.041\\
		& VGG19 & 0.608 $\pm$ 0.161 & 0.954 $\pm$ 0.120\\
		& DenseNet201 & 0.660 $\pm$ 0.194 & 0.978 $\pm$ 0.084\\
		& ResNet152 & \textbf{0.743 $\pm$ 0.163} & \textbf{0.986 $\pm$ 0.089} \\
		& InceptionV3 & 0.561 $\pm$ 0.125 & 0.967 $\pm$ 0.131 \\
		\hline	
	\multirow{5}{*}{Tiger} 
	    & AlexNet & 0.830 $\pm$ 0.296 & 0.978 $\pm$ 0.217\\
		& VGG19 & 0.770 $\pm$ 0.205 & 0.940 $\pm$ 0.145 \\
		& DenseNet201 & \textbf{0.863 $\pm$ 0.193} & 0.974 $\pm$ 0.148 \\
		& ResNet152 & 0.811 $\pm$ 0.124 & \textbf{0.996 $\pm$ 0.072} \\
		& InceptionV3 & 0.731 $\pm$ 0.117 & 0.933 $\pm$ 0.121\\
		\hline	
\end{tabular}
\end{table}

Here we report only the best models per data set. Results for all models can be found in Table 1. 

On the FaceScrub data set, the Triplet Loss DenseNet201 performs the best achieving a 0.914 mAP@1 while ResNet152 achieved the best mAP@5 with 0.952. Considering state-of-the-art models specifically designed for faces, such as FaceNet, which has a re-ID accuracy of 0.9512 on the YouTube Faces dataset, our model does not match their level of performance, indicating there is room for improvement in terms of species-specific architecture \cite{schroff2015facenet}. However, our results do demonstrate the capabilities of this general system for animal re-ID. This serves as platform of comparison for how similarity learning models perform on a variety of different species and data set composition.

On the Chimpface data set, the ResNet152 Triplet Loss network attains a mAP@1 and mAP@5 of 0.811 and 0.961 respectively. This is an improvement over previously reported closed-set rank-1 accuracies \cite{deb2018face} (Figure 4).

On the HappyWhale data set, the Triplet Loss InceptionNetV3 architecture achieved the highest mAP@1 with 0.776 and ResNet152 achieved the highest mAP@5 with 0.860. The lower scores for this data set are likely due to the limitations of animal pairings as this data set has an average number of images per individual at 2.1. This is an improvement over previously reported mAP@5 scores for this dataset of 0.786 \cite{kagglehumpbackreid}

For the FruitFly data set, the Triplet Loss ResNet152 achieved the highest mAP@1 and mAP@5 of 0.743 and 0.989 respectively. Distinguishing between fruit fly individuals seems to be an impossible task for humans, yet our model was able to successfully learn features which accurately distinguish between individuals. Considering the data set, the success of this model is likely based on the very large number of training images available, the limited number of individuals, as well as the standardized background of the images themselves. To our knowledge this is the first re-ID experiment considering this dataset. 

For the Amur Tiger Re-identification in the Wild dataset, the Triplet Loss DenseNet201 achieved the highest mAp@1 score with 0.863, while ResNet152 achieved the highest mAP@5 score of 0.996. This data set achieved the highest mAP@5 scores, likely due to the fact that tigers have the most distinguishable features in comparison to all species considered. To our knowledge this is the first re-ID experiment considering this dataset. 

\section{Near Future Techniques for Animal Re-Identification}

The success of our models across multiple species, phyla, and environments show that similarity learning networks are capable of generalizing across species with high performance and can be used as an approach to address the one-shot learning problem associated with animal population monitoring. Ecologists can realize our suggested similarity networks to improve accuracies of computer vision aided animal re-ID without the requirement of hand-coded feature extraction methods and example images from every member of the population.

Our results were unanimous in that Triplet Loss networks outperform Siamese nets for the five species data sets considered. In terms of network selection, our results found that ResNet152 performed optimally considering the mAP@5 metric for all species. This indicates ResNet152 is best for generalization across species. When considering mAP@1, DenseNet201, InceptionV3, and ResNet152 each had optimal performances. 

To design a system that utilizes this technique, in practice a wildlife re-ID system would work as follows. One would collect or find a data library of images of animal individuals for the species in consideration, which under ideal conditions data should: a) not based on video frame data due to background biases, b) have 500+ individuals and c) has 2-5+ sightings for each. For image libraries with similar backgrounds one can then utilize animal location and background subtraction (MASK-RCNN) to add alternative backgrounds as a form of image augmentation \cite{parham2018animal, schneider2018deep, he2017mask}. One would then train a similarity network. Once train, a similarity network can be used as a method of population estimates by querying a database. Upon initialization, this database would be empty. As each individual enters into the camera, the network would query all existing animal within the database. If none are deemed to be similar, an image of the new individual would be added to the database and the process repeats for each individual that enters. Eventually, this approach would see diminishing returns with increased uncertainty for each individual added. We recommend having multiple examples of each individual to mitigate this problem.  

One of the limitations of this work, and the greatest challenges within the animal re-ID computer vision literature, is data availability. Due to the difficulty of data collection in this domain, we encourage researchers with images of labeled animal individuals to make these data sets publicly available to further the research in this field. 

In order for such a technique to become generally applicable, we foresee the greatest challenge for deep learning methods being the creation of large labeled data sets for animal individuals. Our proposed approach for data collection would be to utilize environments with known ground truths for individuals, such as national parks, zoos, or camera traps in combination with individuals being tracked by GPS, to build the data sets. We recommend using video wherever possible to gather the greatest number of images for a given encounter with an individual, but be careful not to use repetitious images. Due to the difficulty of data collection in this domain, we encourage researchers with images of labeled animal individuals to make these data sets publicly available to further the research in this field. In addition to gathering the images, labeling data is also a labourious task, especially when training an object detection model where bounding boxes are required. One approach for solving this problem is known as weakly supervised learning, where one provides object labels to a network (i.e.~zebra) and the network returns the predicted coordinates of its location \cite{zhou2017brief}. An alternative approach is to outsource the labeling task to online services, such as Zooniverse which can be time saving for researchers, but introduces inevitable variability in the quality of annotations \cite{simpson2014zooniverse}. 

While deep learning approaches are able to generalize to examples similar to those seen during training, we foresee various environmental, positional, and timing related challenges. Environmental challenges may include inclement weather conditions, such as heavy rain, or extreme lighting/shadows, especially from video analysis which only makes comparisons between similar weather conditions. One possible solution to limit these concerns may be to re-ID only during optimal weather conditions. A second is to include a robust amount of image augmentation. A positional challenge may occur if an individual were to enter the camera frame at extremely near or far distances. To solve this, one could limit animals to a certain range from the camera before considering it for re-ID. A challenge may also arise if an individual's appearance were to change dramatically between sightings, such as being injured or the rapid growth of a youth. While a network would be robust to such changes given training examples, this would require examples be available as training data. To account for this, and all the other listed considerations, we would recommend having a `human-in-the-loop' approach, where a human monitors results and relabels erroneous classifications for further training to improve performance \cite{holzinger2016interactive}.

An area of future research for animal re-ID is to test how do accuracies change as we increase and decrease the number of sample images per individual. This includes if there is a large class imbalance, where few individuals dominate the majority of images, as well as if there are limited numbers of images throughout (i.e. only 2-3 images of each individual). Additionally, one could compare and contrast the re-ID accuracy per species when training data are similar by investigating how distinguishable markings, such as strips, improve performance in comparison to species that are harder to distinguish.

While today fully autonomous animal re-ID is still in early stages, ecologists can already use machine learning systems to reduce manual labour for their studies. Examples include training networks to filter images by the presence/absence of animals, species classifications, and/or object detection methods \cite{kaggleiWildCam2018, norouzzadehautomatically, schneider2018deep}. If the current rate of advancement continues, soon deep learning systems will accurately perform animal re-ID in real-world environmental conditions. At this time one can create systems that autonomously extract from camera traps a variety of ecological metrics such as diversity, evenness, richness, relative abundance distribution, carrying capacity, and trophic function, contributing to overarching ecological interpretations of trophic interactions and population dynamics. This will allow us to receive autonomous updates of projected ecosystem and species population health.

\section{Conclusion}

The ability for a researcher to re-identify an animal individual upon re-encounter is fundamental for addressing a broad range of questions in the study of population dynamics and community/behavioural ecology. Standard deep learning methodologies are not viable for this task as gathering a library of images of all animal individuals from a population is infeasible. One-shot learning for re-ID shows promise to solve the animal re-ID task as only one image is required. We tested the capabilities of two similarity comparison frameworks, Siamese and Triplet Loss networks using AlexNet, VGG19, DenseNet201, ResNet152, and InceptionV3 architectures considering closed-set re-ID as a five-fold split on five different species data sets: humans, chimpanzees, humpback whales, fruit fly and Siberian tigers. 

Our results find that unanimously Triplet Loss outperformed Siamese networks for all species. Our results also found that ResNet152 obtained the highest mAP@5 score across species, while DenseNet201, ResNet152, and InceptionV3 each performed best for mAP@1 scores for humans, chimpanzees and fruit flies, humpback whale and tiger respectively. Our results on fruit flies demonstrate that similarity comparison networks can achieve accuracies beyond human level performance for the task of animal re-ID. Our expectation is that similarity comparison networks are the beginning of a major trend that could stand to revolutionize the analysis of animal re-ID camera trap data and, ultimately, our approach to animal ecology.

{
\small
\bibliographystyle{IEEEtranN}
\bibliography{main}
}

\end{document}